\documentclass[runningheads,a4paper]{llncs}
\usepackage[T1]{fontenc}
\usepackage{graphicx}
\usepackage{amsmath,amssymb}
\usepackage{booktabs}
\usepackage{url}

\begin{document}
\title{Finding Change in Satellite Archives from Text: How to Combine Before-and-After Images Efficiently}
\titlerunning{Bi-Temporal Fusion for Text-Based Change Retrieval}
\author{Simon Roy \and
Haechan Mark Bong \and
Giovanni Beltrame}
\authorrunning{S. Roy et al.}
\institute{Polytechnique Montr\'eal, Montr\'eal, QC, Canada\\
\email{\{simon-7.roy,haechan.bong,giovanni.beltrame\}@polymtl.ca}}
\maketitle
\begin{abstract}
Operational Earth observation increasingly calls for answering queries such as
``find the image pairs where a new building appeared.'' This means searching
an archive of before-and-after (bi-temporal) satellite image pairs and ranking
each pair by how well it matches a natural-language description of the change.
The component that performs this match, the fusion module that combines the
``before'' and ``after'' views, must be run at query time across many
candidate pairs, so its speed largely sets the cost of every search. We
present a controlled comparison of how to build that module. Using one fixed
image encoder (a frozen CLIP model) and one training recipe for all variants,
we evaluate eight designs drawn from three families: attention, state-space
models (Mamba), and learned compression (our Temporal Bottleneck Fusion, TBF).
Each design is tested on two benchmarks (LEVIR-CC and Dubai-CC) with ten
random seeds, so the reported differences are statistically grounded. We
outline three findings: first, a training-free two-stage search (a cheap
difference model that shortlists candidates, followed by attention fusion that
re-ranks them) matches or exceeds full-fusion recall on LEVIR-CC while cutting
query cost $10$-$15\times$, with comparable R@1/R@5 on Dubai-CC; second, the linear-time
scan of Mamba, attractive on paper, gives no speed benefit at the patch counts
typical of vision transformers ($L{=}196$): the scan is limited by memory
bandwidth, whereas attention maps cleanly onto parallel hardware; and third,
compressing the fused representation (TBF) reduces parameters by $2.3\times$
and latency by $1.6\times$ for a change-only BLEU-1 cost of $0.007$,
although more aggressive compression quietly discards change-relevant detail
that aggregate metrics fail to reveal.

\end{abstract}
\begin{figure}[t]
    \centering
    \includegraphics[width=1\linewidth]{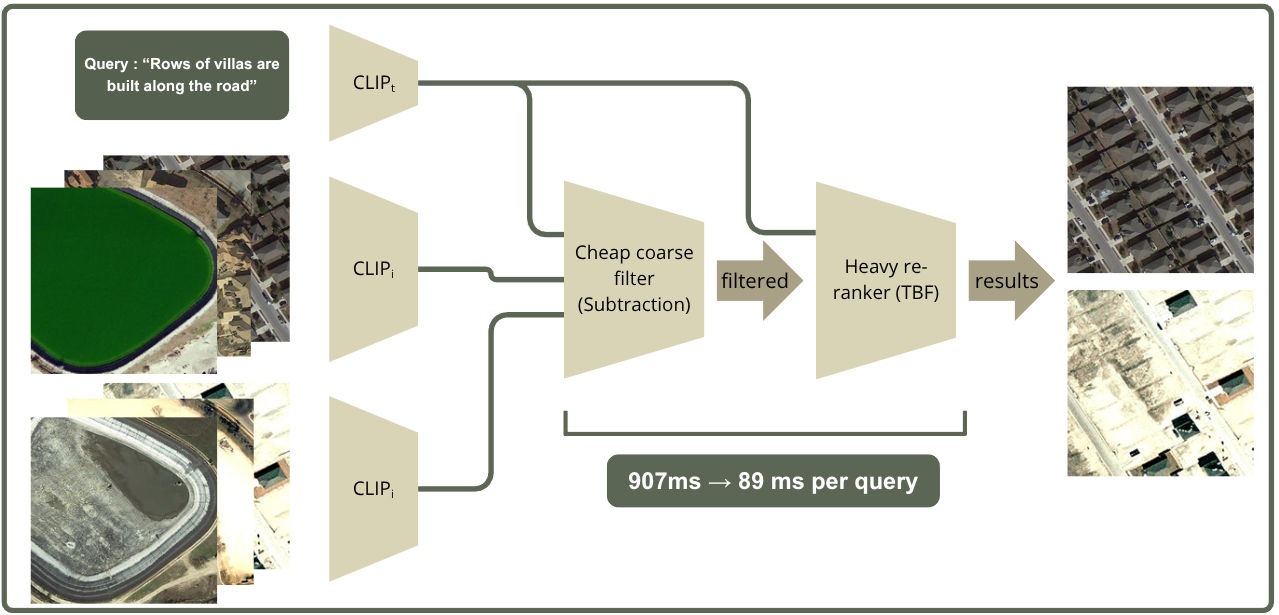}
    \caption{Cascaded change retrieval (LEVIR-CC, $G{=}1929$, $N{=}25$). Frames are CLIP-encoded offline; a subtraction filter ranks all candidate pairs, then TBF re-ranks the top-$N$ shortlist, cutting fusion latency from 907 to 89\,ms per query while matching or exceeding full-fusion recall.}
    \label{fig:pipeline}
\end{figure}
\section{Introduction}
\label{sec:intro}

Earth observation (EO) archives are growing at a pace that rules out manual analysis, and many of the most valuable insights are not visible in any single image. Assessing storm damage or auditing urban expansion requires comparing two acquisitions, e.g., finding every tile where ``a building appears along the road''. Text-based change retrieval answers such queries directly, given a natural-language description of a change, it returns the bi-temporal image pairs that exhibit it, replacing the manual screening of thousands of candidates.

Vision-language models such as CLIP~\cite{radford2021learningtransferablevisualmodels} embed images and text in a shared latent space and have been adapted to remote sensing~\cite{liu2024remoteclipvisionlanguagefoundation}. Text-ITSR~\cite{hoxha2025selfsupervisedcrossmodaltextimagetime} extends the paradigm to bi-temporal retrieval by fusing the features of the two acquisitions with a Transformer module before matching against the text embedding. In a realistic deployment, the expensive vision backbone runs once per frame at ingestion, while the fusion module runs at query time over every candidate pair formed by the requested region and time interval (Fig.~\ref{fig:pipeline}, Sect.~\ref{sec:deployment}): fusion latency and memory govern interactive query cost, especially on constrained platforms.

Two separate factors drive the cost of Transformer-based fusion. The first is the number of image patches $L$ that the model attends over: self-attention compares every patch with every other, so its cost grows with $L^2$. The second is the feature width $D$, the length of the vector used to describe each patch. Stitching the ``before'' and ``after'' views together doubles this width, and because the cost of the model's linear and feed-forward layers grows with $D^2$, doubling $D$ roughly quadruples their cost. The two factors are independent: one is set by image resolution, the other by how the two views are combined. At standard ViT patch counts ($L=196$), the width factor dominates the parameter budget. The two factors call for different remedies: state space models (SSMs) such as Mamba~\cite{gu2024mambalineartimesequencemodeling} attack the sequence-length term with linear-time selective scans, whereas compression schemes attack the width term by reducing $D$ before interaction.

Rather than proposing another fusion architecture, we ask which fusion family offers the best efficiency-accuracy trade-off for text-based change retrieval, and whether theoretical complexity advantages show up in wall-clock benchmarks. We compare eight fusion designs spanning attention-based (concatenation Transformer, cross-attention transformer feature fusion (TFF)\cite{hoxha2025selfsupervisedcrossmodaltextimagetime}), SSM-based (three Mamba variants), and compression-based (Temporal Bottleneck Fusion, TBF, applying the classical concat-then-reduce pattern before self-attention) families, plus two simple baselines, under identical backbone, data, and training conditions on LEVIR-CC and Dubai-CC. Our contributions are:
\begin{itemize}
    \item a training-free retrieval cascade that yields $10$-$15\times$ query-cost reductions while preserving full-fusion recall.
    \item to our knowledge the first controlled comparison of attention, bottleneck, and SSM-based bi-temporal fusion for text-based change retrieval, eight designs under one backbone and training recipe with statistical reporting over ten seeds.
    \item a practical analysis showing that Mamba's linear-time advantage does not translate into latency benefits at standard ViT sequence lengths ($L{=}196$), where its memory-bound scan trails parallel attention.
    \item the identification of TBF as a favorable efficiency-quality operating point, cutting parameters $2.3\times$ and latency $1.6\times$ for $0.007$ change-only BLEU-1.
\end{itemize}

\section{Related Work}
\label{sec:related}

\paragraph{Cross-Modal Retrieval and Change Understanding in EO.}
Text-based retrieval projects images and language into a shared latent space via contrastive frameworks such as CLIP~\cite{radford2021learningtransferablevisualmodels}. Remote-sensing models such as RemoteCLIP~\cite{liu2024remoteclipvisionlanguagefoundation} excel at single-image tasks but lack bi-temporal mechanisms. Change detection has evolved from Siamese networks~\cite{daudt2018fullyconvolutionalsiamesenetworks} to Transformers such as BIT~\cite{9491802}, and change captioning~\cite{9934924} extends supervision to natural language. Text-based change retrieval itself is nascent, Ferrod et al.~\cite{ferrod2024multimodalframeworkremotesensing} first aligned bi-temporal change embeddings with text queries on LEVIR-CC, and Text-ITSR~\cite{hoxha2025selfsupervisedcrossmodaltextimagetime} combines a Siamese encoder with a Transformer fusion stage. Both treat temporal interaction as a generic sequence problem, computational trade-offs of the fusion stage are not evaluated in this literature.

\paragraph{State Space Models.}
To address the quadratic complexity of attention, SSMs such as Mamba~\cite{gu2024mambalineartimesequencemodeling} offer linear-time alternatives with promising results in dense remote sensing prediction such as ChangeMamba~\cite{Chen_2024}. Most SSM evidence comes from dense long-sequence tasks ($L > 1024$); vision-language adoption remains recent~\cite{zhao2025cobraextendingmambamultimodal}.

\paragraph{Bottleneck and Early-Fusion Strategies.}
Concat-then-reduce is an established design pattern: FC-Siam-conc~\cite{daudt2018fullyconvolutionalsiamesenetworks} fuses concatenated encoder features with learned convolutions, BIT~\cite{9491802} projects fused features into a compact token space, and the Multimodal Bottleneck Transformer~\cite{nagrani2022attentionbottlenecksmultimodalfusion} restricts information flow for efficient multimodal fusion. The Temporal Bottleneck Fusion (TBF) module evaluated here transfers this concat-then-reduce principle to cross-modal change retrieval.
\section{Study Design}
\label{sec:method}

\subsection{Task Definition and Retrieval Protocol}
\label{sec:protocol}
Let $(\mathcal{I}_1, \mathcal{I}_2)$ be co-registered images at times $t_1, t_2$ and $\mathcal{T}$ a natural-language description, the goal is an alignment score $S(\mathcal{I}_1, \mathcal{I}_2, \mathcal{T})$ ranking correct pairs higher. Every model uses the same frozen CLIP ViT-B/16 encoder to extract patch features $X_1, X_2 \in \mathbb{R}^{L \times D_{in}}$ ($L=196$, $D_{in}=768$), projected to width $D=320$ before fusion. Similarity between fused visual embedding $v_i$ and text embedding $t_j$ is the cosine score
\begin{equation}
  s_{ij}= \frac{v_i^{\top} t_j}{\lVert v_i \rVert_2\, \lVert t_j \rVert_2},
\end{equation}
and all models are trained with the standard symmetric InfoNCE objective over $s_{ij}$.

\paragraph{Ground Truth and Matching.}
    Each pair carries five captions. A query is a single caption whose unique ground truth is the pair it annotates (captions are not clustered across pairs). Recall@$K$ therefore measures strict instance-level success. Because archives contain many similar scenes undergoing similar changes, strict instance matching underestimates practical utility. We therefore complement Recall@$K$ with semantic metrics comparing the query caption against the reference captions of the top-1 retrieved pair (BLEU-1/4, METEOR, ROUGE-L~\cite{sharma2017nlgeval}), a retrieval returning a different pair with the same change still scores well.

\subsection{Datasets}
\textbf{LEVIR-CC}~\cite{9934924} contains high-resolution (0.5\,m/px) RGB pairs over Texas in $256 \times 256$ patches, focused on urbanization, five captions per pair, our split holds 3,918/1,333/1,929 pairs (train/val/test, 70/30 change ratio pooled; test is balanced, 964 change pairs). Following~\cite{hoxha2025selfsupervisedcrossmodaltextimagetime}, the split is retrieval-oriented (more change pairs than the standard captioning split) and applies identically to all compared models. \textbf{Dubai-CC}~\cite{hoxha2022changecaptioning} provides similar annotations on $50{\times}50$ Landsat\,7 tiles at ${\sim}30$\,m/px, standard split (300/50/150, 65/35 change ratio).

\subsection{Fusion Architectures Under Study}
\label{sec:architectures}
All fusion modules consume the projected features and produce a single visual embedding, everything else is held fixed.

\paragraph{Simple Baselines.}
\emph{Subtraction} computes the element-wise difference $X_2 - X_1$ followed by a linear projection. MLP Fusion concatenates the two streams channel-wise and applies a two-layer MLP, i.e., channel mixing without token interaction.

\paragraph{Attention-Based.}
The Concatenation Transformer concatenates features along the channel axis ($L \times 2D$) and processes them with a standard Transformer encoder, the doubled width inflates every linear projection and feed-forward block, yielding 13.05M parameters. TFF, the fusion stage of Text-ITSR~\cite{hoxha2025selfsupervisedcrossmodaltextimagetime}, applies cross-attention between difference features and the original bi-temporal features.

\paragraph{Compression-Based: Temporal Bottleneck Fusion (TBF).}
TBF applies the concat-then-reduce pattern before attention: (i) compression: an MLP (Linear $2D{\to}D$, ReLU, Linear $D{\to}D$, LayerNorm) projects the concatenated features back to width $D$, (ii) interaction: a standard Transformer encoder operates at width $D$ instead of $2D$. Since the cost of projections and feed-forward blocks scales quadratically with width, operating at $D$ rather than $2D$ reduces their parameters and computation by roughly $4\times$, while preserving global self-attention across all patches.

\paragraph{SSM-Based.}
We evaluate three Mamba~\cite{gu2024mambalineartimesequencemodeling} variants inspired by recent change-detection work~\cite{Chen_2024}. Concatenation Mamba replaces the Transformer of the concatenation baseline with Mamba blocks at width $2D$. Bottleneck Mamba uses the same compression MLP as TBF followed by Mamba blocks at width $D$. Interleaved Mamba interleaves temporal tokens into a single sequence of length $2L$.

\subsection{Deployment Setting and Implementation}
\label{sec:deployment}
\emph{Offline}, each frame is encoded once at ingestion and stored with metadata; pairs are not precomputed. \emph{Online}, a text query with region/time constraints is matched live against satellite or drone imagery. The number of scanned pairs scales with archive size, while backbone cost is amortized at ingestion, so query latency is governed by the fusion module.

Models are implemented in PyTorch (\texttt{mamba\_ssm}~\cite{gu2024mambalineartimesequencemodeling}) and trained on one RTX 4070. Optimization uses AdamW (learning rate $8\times10^{-5}$, weight decay $5\times10^{-4}$), cosine annealing to $10^{-6}$, batch size 32, dropout 0.25, and 30 epochs. All fusion modules use 3 layers with $d_{model}=320$, attention-based models use 16 heads, SSM-based models use state dimension 16, convolution width 4, and expansion factor 2. Results are mean $\pm$ std over 10 training runs unless stated otherwise. Code and split indices: \url{https://github.com/SimonR99/bitemporal-fusion-benchmark}.

\section{Results}
\label{sec:results}

\subsection{Efficiency and Complexity}
Table~\ref{tab:efficiency_main} compares parameters, FLOPs, and measured latency. Latency is wall-clock under CUDA synchronization (1{,}000 passes after 50 warm-ups, RTX 4070), FLOPs are computed with \texttt{fvcore}, Mamba FLOPs analytically via the selective-scan formula~\cite{gu2024mambalineartimesequencemodeling}.

\begin{table}[t]
    \centering
    \caption{Efficiency of fusion mechanisms at $L=196$. Latency is a single fused forward pass (batch 1), averaged over 1{,}000 runs on an RTX 4070. Bold = best among the six learned modules.}
    \label{tab:efficiency_main}
    \setlength{\tabcolsep}{8pt}
    \footnotesize
    \begin{tabular}{l c c c}
        \toprule
        \textbf{Model} & \textbf{Params (M)} & \textbf{FLOPs (G)} & \textbf{Latency (ms)} \\
        \midrule
        Subtraction & 0.35 & 0.14 & 0.04 \\
        MLP Fusion & 2.71 & 1.06 & 0.20 \\
        \midrule
        Concat.\ Transformer & 13.05 & 5.42 & 0.76 \\
        TFF (Text-ITSR) & 8.02 & 2.65 & 1.90 \\
        TBF & 5.73 & 2.41 & \textbf{0.47} \\
        \midrule
        Concat.\ Mamba & 8.14 & 3.28 & 0.58 \\
        Bottleneck Mamba & 2.58 & \textbf{1.06} & 0.62 \\
        Interleaved Mamba & \textbf{2.27} & 1.77 & 0.60 \\
        \bottomrule
    \end{tabular}
\end{table}

\paragraph{Bottleneck Compression.}
TBF uses $2.3\times$ fewer parameters than the Concatenation Transformer (5.73M vs.\ 13.05M) and is $1.6\times$ faster (0.47 vs.\ 0.76\,ms): halving the width before attention removes most of the concatenation cost. A width sweep confirms this $2\times$ bottleneck ($b{=}320$) matches the uncompressed width within noise, while more aggressive compression degrades change-only quality before full-set quality.

\begin{figure}[t]
    \centering
    \begin{minipage}[c]{0.47\linewidth}
        \centering
        \includegraphics[width=\linewidth]{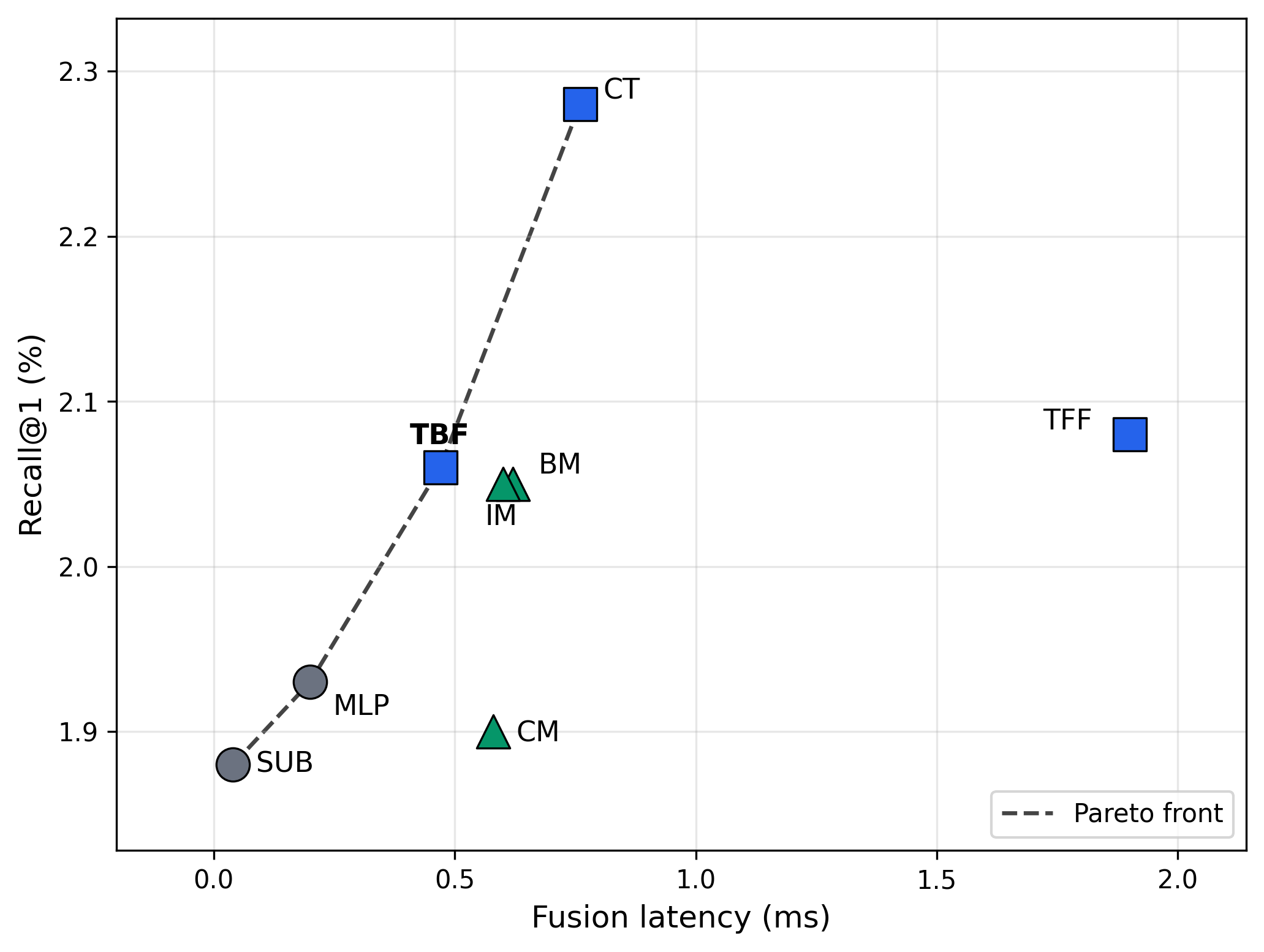}\\
        {\footnotesize (a)}
    \end{minipage}\hfill
    \begin{minipage}[c]{0.47\linewidth}
        \centering
        \includegraphics[width=\linewidth]{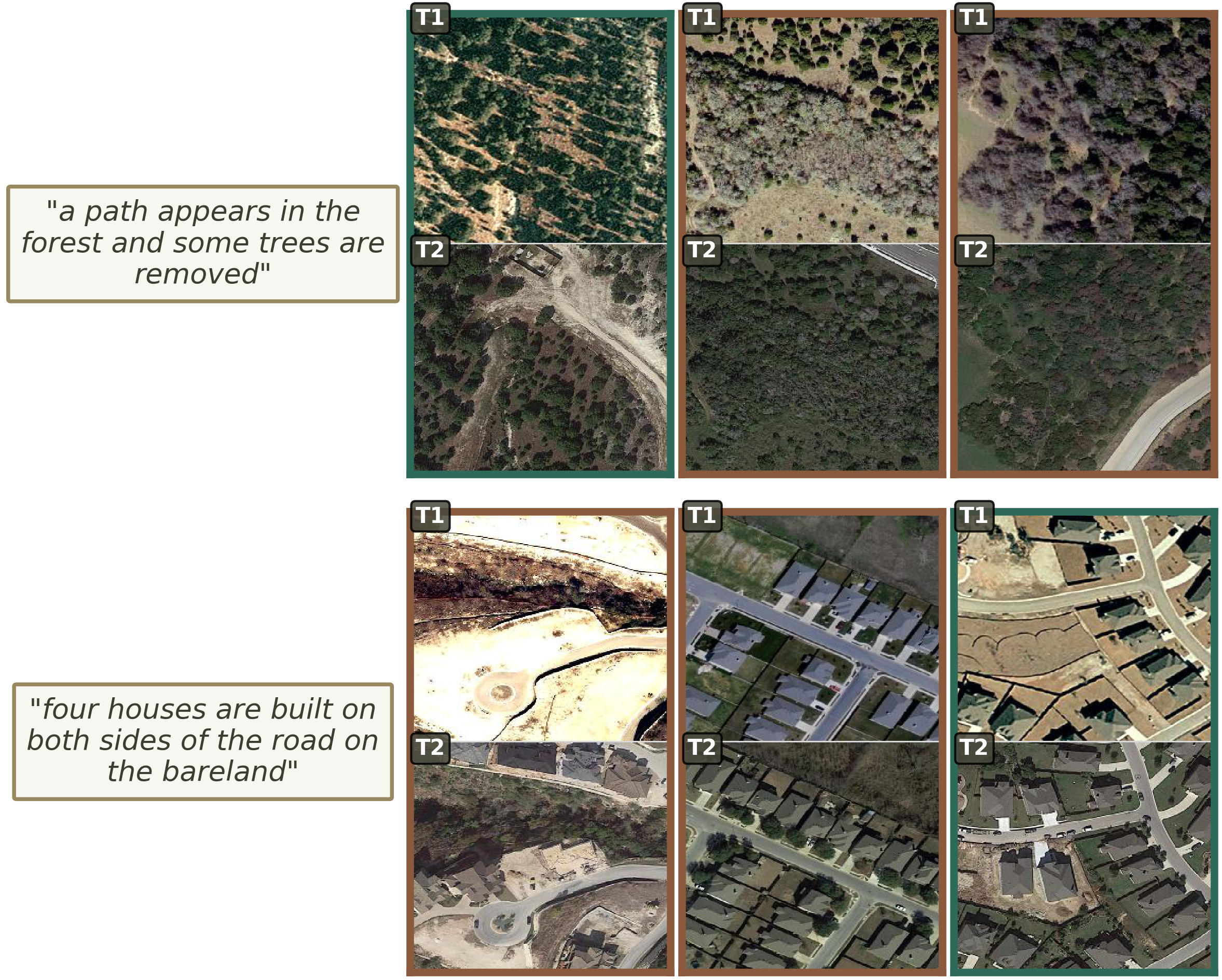}\\
        {\footnotesize (b)}
    \end{minipage}
    \caption{(a)~Recall@1 vs.\ fusion latency on LEVIR-CC change queries (CT = Concatenation Transformer; dashed = Pareto front). (b)~TBF top-3 retrievals; the correct pair (green) ranks first (top) and third (bottom), and the distractors (brown) are semantically plausible.}
    \label{fig:summary}
\end{figure}

\paragraph{SSM Latency.}
Although Interleaved Mamba has the fewest parameters (2.27M) and low FLOPs (1.77G), its latency (0.60\,ms) is 28\% higher than TBF and only 21\% lower than the much larger Concatenation Transformer: theoretical linear complexity does not translate into wall-clock gains (hardware causes: Sect.~\ref{sec:discussion}). Figure~\ref{fig:summary}a plots Recall@1 against latency: the Pareto front contains only the simple baselines, TBF, and the Concatenation Transformer.

\subsection{Semantic Retrieval Quality}
Table~\ref{tab:captioning} reports caption-similarity metrics on the complete test sets and on the change-only subsets, which isolate the matching of actual state transitions rather than background priors.

\begin{table}[t]
    \centering
    \caption{Caption-similarity metrics on the full test sets and change-only subsets (mean $\pm$ std over 10 runs). B-1/B-4 = BLEU-1/4, MET = METEOR, R-L = ROUGE-L. Best per dataset and regime in bold; scores average both retrieval directions.}
    \label{tab:captioning}
    \setlength{\tabcolsep}{3pt}
    \resizebox{\textwidth}{!}{
    \begin{tabular}{l cccc cccc}
        \toprule
        & \multicolumn{4}{c}{\textbf{LEVIR-CC}} & \multicolumn{4}{c}{\textbf{Dubai-CC}} \\
        \cmidrule(lr){2-5}\cmidrule(lr){6-9}
        \textbf{Model} & \textbf{B-1} & \textbf{B-4} & \textbf{MET} & \textbf{R-L} & \textbf{B-1} & \textbf{B-4} & \textbf{MET} & \textbf{R-L} \\
        \midrule
        \multicolumn{9}{l}{\emph{Full set}}\\
        \midrule
        Subtraction & $0.631 \pm 0.007$ & $0.326 \pm 0.004$ & $0.294 \pm 0.004$ & $0.573 \pm 0.008$ & $0.575 \pm 0.011$ & $0.270 \pm 0.014$ & $0.274 \pm 0.008$ & $0.514 \pm 0.011$ \\
        MLP Fusion & $0.658 \pm 0.020$ & $0.352 \pm 0.023$ & $0.307 \pm 0.012$ & $0.610 \pm 0.026$ & $0.576 \pm 0.010$ & $0.268 \pm 0.010$ & $0.266 \pm 0.009$ & $0.500 \pm 0.015$ \\
        Concat.\ Transformer & $\mathbf{0.684} \pm 0.017$ & $\mathbf{0.377} \pm 0.018$ & $\mathbf{0.323} \pm 0.012$ & $\mathbf{0.638} \pm 0.027$ & $0.581 \pm 0.012$ & $0.279 \pm 0.010$ & $0.276 \pm 0.009$ & $0.512 \pm 0.017$ \\
        TFF (Text-ITSR) & $0.662 \pm 0.025$ & $0.353 \pm 0.029$ & $0.310 \pm 0.024$ & $0.617 \pm 0.040$ & $0.581 \pm 0.024$ & $0.276 \pm 0.022$ & $0.271 \pm 0.015$ & $0.508 \pm 0.030$ \\
        TBF & $0.662 \pm 0.030$ & $0.357 \pm 0.030$ & $0.308 \pm 0.018$ & $0.610 \pm 0.043$ & $0.573 \pm 0.011$ & $0.268 \pm 0.010$ & $0.270 \pm 0.009$ & $0.507 \pm 0.015$ \\
        Concat.\ Mamba & $0.661 \pm 0.017$ & $0.355 \pm 0.017$ & $0.309 \pm 0.013$ & $0.612 \pm 0.028$ & $\mathbf{0.585} \pm 0.018$ & $\mathbf{0.280} \pm 0.017$ & $\mathbf{0.277} \pm 0.011$ & $\mathbf{0.518} \pm 0.020$ \\
        Bottleneck Mamba & $0.656 \pm 0.014$ & $0.354 \pm 0.012$ & $0.308 \pm 0.006$ & $0.607 \pm 0.018$ & $0.573 \pm 0.012$ & $0.274 \pm 0.011$ & $0.270 \pm 0.009$ & $0.505 \pm 0.014$ \\
        Interleaved Mamba & $0.648 \pm 0.014$ & $0.344 \pm 0.013$ & $0.305 \pm 0.008$ & $0.602 \pm 0.019$ & $0.579 \pm 0.011$ & $0.273 \pm 0.016$ & $0.275 \pm 0.010$ & $0.514 \pm 0.014$ \\
        \midrule
        \multicolumn{9}{l}{\emph{Change-only subset}}\\
        \midrule
        Subtraction & $0.640 \pm 0.003$ & $0.249 \pm 0.002$ & $0.251 \pm 0.002$ & $0.452 \pm 0.003$ & $0.574 \pm 0.013$ & $0.254 \pm 0.014$ & $0.259 \pm 0.009$ & $0.471 \pm 0.017$ \\
        MLP Fusion & $0.640 \pm 0.009$ & $0.250 \pm 0.012$ & $0.249 \pm 0.006$ & $0.449 \pm 0.009$ & $0.584 \pm 0.011$ & $0.261 \pm 0.013$ & $0.254 \pm 0.007$ & $0.472 \pm 0.014$ \\
        Concat.\ Transformer & $\mathbf{0.655} \pm 0.004$ & $\mathbf{0.267} \pm 0.003$ & $\mathbf{0.259} \pm 0.002$ & $\mathbf{0.464} \pm 0.004$ & $0.585 \pm 0.015$ & $0.270 \pm 0.015$ & $0.263 \pm 0.007$ & $0.483 \pm 0.016$ \\
        TFF (Text-ITSR) & $0.640 \pm 0.007$ & $0.255 \pm 0.006$ & $0.254 \pm 0.003$ & $0.452 \pm 0.004$ & $0.580 \pm 0.022$ & $0.263 \pm 0.020$ & $0.258 \pm 0.010$ & $0.469 \pm 0.023$ \\
        TBF & $0.648 \pm 0.009$ & $0.259 \pm 0.010$ & $0.255 \pm 0.005$ & $0.459 \pm 0.010$ & $0.576 \pm 0.010$ & $0.257 \pm 0.009$ & $0.256 \pm 0.006$ & $0.469 \pm 0.009$ \\
        Concat.\ Mamba & $0.645 \pm 0.003$ & $0.255 \pm 0.006$ & $0.251 \pm 0.003$ & $0.453 \pm 0.003$ & $\mathbf{0.592} \pm 0.013$ & $\mathbf{0.272} \pm 0.017$ & $\mathbf{0.263} \pm 0.011$ & $\mathbf{0.484} \pm 0.014$ \\
        Bottleneck Mamba & $0.643 \pm 0.003$ & $0.255 \pm 0.004$ & $0.252 \pm 0.003$ & $0.452 \pm 0.004$ & $0.577 \pm 0.014$ & $0.263 \pm 0.014$ & $0.256 \pm 0.008$ & $0.467 \pm 0.012$ \\
        Interleaved Mamba & $0.639 \pm 0.005$ & $0.253 \pm 0.006$ & $0.252 \pm 0.003$ & $0.453 \pm 0.005$ & $0.586 \pm 0.013$ & $0.262 \pm 0.022$ & $0.260 \pm 0.011$ & $0.475 \pm 0.016$ \\
        \bottomrule
    \end{tabular}
    }
\end{table}

On LEVIR-CC the Concatenation Transformer is consistently best, with TBF the closest competitor at $2.3\times$ fewer parameters and $1.6\times$ lower latency. Welch tests ($n{=}10$, change-only BLEU-1) separate CT from six of the seven modules ($p{\leq}0.0002$). TBF is the exception: its gap of $0.007$ is $1.6\times$ smaller than the next closest and sits at the edge of detectability, significant unpaired ($p{=}0.04$) but not paired per seed ($p{=}0.08$). On Dubai-CC the modules cluster within noise (change-only BLEU-1 within $0.02$), with 300 training pairs, higher-capacity fusion cannot be exploited (Sect.~\ref{sec:discussion}).

\subsection{Instance-Level Retrieval Accuracy}
Table~\ref{tab:retrieval_change} reports strict Recall@$K$ for change-caption queries ranked against the \emph{full} test gallery (the random baseline reflects the gallery size); no-change queries are excluded as uninformative for instance ranking.

\begin{table}[t]
    \centering
    \caption{Strict instance-level retrieval accuracy (Recall@$K$, \%) for change-caption queries over the full test gallery (deduplicated to unique image pairs; mean $\pm$ std over 10 runs). Best per dataset in bold.}
    \label{tab:retrieval_change}
    \setlength{\tabcolsep}{5pt}
    \resizebox{\textwidth}{!}{
    \begin{tabular}{l c c c c c c}
        \toprule
        & \multicolumn{3}{c}{\textbf{LEVIR-CC (Change Queries)}} & \multicolumn{3}{c}{\textbf{Dubai-CC (Change Queries)}} \\
        \cmidrule(lr){2-4}\cmidrule(lr){5-7}
        \textbf{Model} & \textbf{R@1} & \textbf{R@5} & \textbf{R@10} & \textbf{R@1} & \textbf{R@5} & \textbf{R@10} \\
        \midrule
        Random & $0.05$ & $0.26$ & $0.52$ & $0.67$ & $3.33$ & $6.67$ \\
        Subtraction & $1.88 \pm 0.17$ & $7.93 \pm 0.25$ & $13.66 \pm 0.39$ & $3.11 \pm 0.82$ & $14.70 \pm 2.38$ & $26.33 \pm 2.79$ \\
        MLP Fusion & $1.93 \pm 0.26$ & $7.85 \pm 0.69$ & $13.66 \pm 1.23$ & $4.82 \pm 1.05$ & $19.51 \pm 1.97$ & $33.38 \pm 2.21$ \\
        Concat.\ Transformer & $\mathbf{2.28} \pm 0.26$ & $\mathbf{8.91} \pm 0.81$ & $\mathbf{15.27} \pm 1.15$ & $5.20 \pm 1.02$ & $21.01 \pm 1.70$ & $35.81 \pm 2.20$ \\
        TFF (Text-ITSR) & $2.08 \pm 0.22$ & $8.00 \pm 0.40$ & $13.77 \pm 0.58$ & $4.21 \pm 0.70$ & $18.97 \pm 2.56$ & $33.44 \pm 3.38$ \\
        TBF & $2.06 \pm 0.42$ & $8.29 \pm 0.94$ & $14.41 \pm 1.39$ & $4.82 \pm 1.17$ & $19.73 \pm 1.63$ & $33.63 \pm 1.74$ \\
        Concat.\ Mamba & $1.90 \pm 0.23$ & $7.80 \pm 0.43$ & $13.38 \pm 0.69$ & $\mathbf{5.32} \pm 1.61$ & $\mathbf{22.10} \pm 2.09$ & $\mathbf{36.33} \pm 2.02$ \\
        Bottleneck Mamba & $2.05 \pm 0.15$ & $8.32 \pm 0.27$ & $14.33 \pm 0.40$ & $4.97 \pm 0.55$ & $19.20 \pm 2.05$ & $33.07 \pm 2.27$ \\
        Interleaved Mamba & $2.05 \pm 0.27$ & $8.07 \pm 0.55$ & $13.75 \pm 0.74$ & $4.58 \pm 1.16$ & $19.63 \pm 1.64$ & $32.54 \pm 1.33$ \\
        \bottomrule
    \end{tabular}
    }
\end{table}

The Concatenation Transformer leads on LEVIR-CC, with TBF competitive ($8.29$/$14.41$ at R@5/R@10 vs.\ $8.91$/$15.27$; Welch $p{\geq}0.13$). On the much smaller Dubai-CC the modules are noise-dominated (R@1 spans $4.2$--$5.3$, std ${\sim}1$; Subtraction lowest): the nominal Dubai leader (Concat.\ Mamba) is near-bottom on LEVIR, so no module separates significantly. LEVIR-CC values are low because several pairs satisfy the same change description; as Fig.~\ref{fig:summary}b shows, incorrect retrievals are typically semantically plausible, so strict Recall@$K$ understates practical retrieval quality.

\subsection{Cascaded Retrieval: Exploiting the Trade-Off}
\label{sec:cascade}

Multi-stage ranking is classical in information retrieval~\cite{wang2011cascade} and standard in vision-language retrieval. Since deployment pays the fusion cost for every candidate pair (Sect.~\ref{sec:deployment}), we test it for change retrieval (Fig.~\ref{fig:pipeline}): stage~1 ranks the full gallery with Subtraction (0.04\,ms/pair), stage~2 re-ranks the top-$N$ with an attention module, reusing the trained models of Table~\ref{tab:efficiency_main} without retraining. Table~\ref{tab:cascade} reports $N{=}25$ against full fusion on LEVIR-CC (deduplicated gallery).

\begin{table}[t]
    \centering
    \caption{Cascaded retrieval on LEVIR-CC (change queries, full gallery $G{=}1929$, mean $\pm$ std over 10 seeds). Per-query cost is the single-stream fusion cost $G\,t_{\mathrm{sub}} + N\,t_{\mathrm{fuse}}$ from Table~\ref{tab:efficiency_main}.}
    \label{tab:cascade}
    \setlength{\tabcolsep}{5pt}
    \scriptsize
    \begin{tabular}{l r c c c r r}
        \toprule
        \textbf{Re-ranker} & \textbf{$N$} & \textbf{R@1} & \textbf{R@5} & \textbf{R@10} & \textbf{ms/query} & \textbf{speedup} \\
        \midrule
        TBF & 25 & $\mathbf{2.33} \pm 0.34$ & $\mathbf{8.91} \pm 0.58$ & $\mathbf{15.17} \pm 0.67$ & 89 & 10.2$\times$ \\
        TBF & full & $2.06 \pm 0.42$ & $8.29 \pm 0.94$ & $14.41 \pm 1.39$ & 907 & 1.0$\times$ \\
        \midrule
        Concat.\ Transf. & 25 & $\mathbf{2.43} \pm 0.31$ & $\mathbf{9.33} \pm 0.53$ & $\mathbf{15.53} \pm 0.53$ & 96 & 15.2$\times$ \\
        Concat.\ Transf. & full & $2.28 \pm 0.26$ & $8.91 \pm 0.81$ & $15.27 \pm 1.15$ & 1466 & 1.0$\times$ \\
        \bottomrule
    \end{tabular}
\end{table}

On LEVIR-CC the cascade is a strict improvement: all evaluated $N$ from 25 to 500 match or exceed full fusion, so no budget tuning is required. In paired per-seed tests at $N{=}25$, the cascade beats full fusion on all ten seeds for TBF at R@1/R@5 ($p{=}10^{-4}$, $0.002$) and on 8 of 10 for the Concatenation Transformer ($p{\leq}0.03$, R@10 not significant). A plausible explanation is an implicit ensemble effect: a candidate must score well under both the difference signal and full fusion, so the prefilter discards distractors that fusion alone ranks highly. On Dubai-CC (retrained from scratch, 10 seeds, $G{=}150$) the gain does not replicate but quality parity does: at $N{=}50$ the cascade matches full fusion on R@1/R@5 with about one point lower R@10, at $2.6\times$ lower cost. Because stage~1 is $12$-$19\times$ cheaper per pair, the cascade speedup grows with gallery size toward this per-pair ceiling: $10$-$15\times$ at $G{=}1929$ and $12$-$19\times$ at $G{=}10^{5}$.

\subsection{Backbone Robustness}
To rule out a CLIP-specific artifact, we repeat the full eight-module comparison with two alternative frozen encoders, \textbf{GeoRSCLIP}~\cite{zhang2024georsclip}, a remote-sensing ViT-B/32, and \textbf{SigLIP\,2}~\cite{tschannen2025siglip2}, a general ViT-B/16 at 256\,px. The fusion ranking is broadly preserved (Spearman $\rho{=}0.69$ and $0.74$ against the CLIP ordering on change-only BLEU-1). Subtraction stays weakest, the attention modules and TBF strongest, and no SSM variant overtakes attention. CLIP ViT-B/16 and SigLIP\,2 lead in absolute terms (change-only BLEU-1 $0.655$ and $0.652$ vs.\ $0.581$), but the relative conclusions do not depend on the backbone.

\section{Discussion}
\label{sec:discussion}

\paragraph{Why Mamba Does Not Pay Off at $L=196$.}
The bottleneck is hardware utilization, the selective scan is sequential and memory-bound while self-attention maps onto highly parallel tensor-core matrix multiplications, consistent with the analysis that motivated Mamba-2~\cite{mamba2}. A sequence-length sweep confirms this, Mamba overtakes attention only beyond $L{\approx}400$ (two frames) and would be $10\times$ faster by $L{=}6272$, so SSMs pay off for long multi-temporal stacks, not at standard tile sizes.

\paragraph{What Bottleneck Compression Buys, and What It Costs.}
TBF never numerically beats the Concatenation Transformer, and the residual change-only gap is small ($0.007$ BLEU-1; Sect.~\ref{sec:results}), while the efficiency savings are deterministic and compound with the number of scanned pairs (Sect.~\ref{sec:deployment}). We hypothesize the gap stays small because bi-temporal scenes are mostly unchanged, so compression loses little~\cite{nagrani2022attentionbottlenecksmultimodalfusion}.

\paragraph{Cross-Dataset Differences and the Low-Data Regime.}
Captioning scores are lower on Dubai-CC while Recall@$K$ is higher (small retrieval pool), with only 300 training pairs, rankings compress and variance is high, so we treat its accuracy as indicative and use it to confirm the efficiency ranking transfers.

\paragraph{Limitations.}
(i)~Our retrieval-oriented LEVIR-CC split prevents comparison with published change-captioning numbers. (ii)~The $n$-gram metrics penalize valid paraphrases and only proxy retrieval utility. (iii)~Latency is measured on one desktop GPU.

\section{Conclusion}
Our controlled study yields three guidelines for text-based change retrieval: at standard patch counts, linear-complexity SSM fusion gives no wall-clock benefit, so optimized attention should remain the default, bottleneck compression before attention (TBF) cuts parameters $2.3\times$ and latency $1.6\times$ for $0.007$ change-only BLEU-1 and fusion need not run on every pair, since a subtraction-prefiltered cascade preserves or improves recall at $10$-$15\times$ lower query cost.

%
\bibliographystyle{splncs04}
\bibliography{main}
\end{document}